\documentclass[letterpaper,11pt]{article}
\usepackage{graphicx}
\usepackage{amsmath}
\usepackage{amssymb}
\usepackage[margin=1.25in]{geometry}
\usepackage[hidelinks,pdfstartview=FitH]{hyperref}
\usepackage[authoryear]{natbib}

\newcommand{\E}{{\bf E}}

\title{Poor starting points in machine learning}
\author{Mark Tygert\\\ \\
Facebook Artificial Intelligence Research\\
1 Facebook Way, Menlo Park, CA 94025\\\ \\tygert@fb.com or tygert@aya.yale.edu}

\begin{document}

\maketitle

\begin{abstract}
Poor (even random) starting points for learning/training/optimization
are common in machine learning.
In many settings, the method of Robbins and Monro
(online stochastic gradient descent)
is known to be optimal for good starting points,
but may not be optimal for poor starting points --- indeed,
for poor starting points Nesterov acceleration can help
during the initial iterations, even though Nesterov methods
not designed for stochastic approximation could hurt during later iterations.
A good option is to roll off Nesterov acceleration for later iterations.
The common practice of training with nontrivial minibatches
enhances the advantage of Nesterov acceleration.
\end{abstract}

\section{Introduction}

The scheme of Robbins and Monro (online stochastic gradient descent)
has long been known to be optimal for stochastic approximation/optimization
\dots\,provided that the starting point for the iterations is good.
Yet poor starting points are common in machine learning,
so higher-order methods (such as Nesterov acceleration) may help
during the early iterations (even though higher-order methods generally hurt
during later iterations), as explained below.
Below, we elaborate on Section~2 of~\citet{sutskever-martens-dahl-hinton},
giving some more rigorous mathematical details, somewhat similar to those given
by~\citet{defossez-bach} and~\citet{flammarion-bach}.
In particular, we discuss the key role of the noise level of estimates
of the objective function, and stress that a large minibatch size
makes higher-order methods more effective for more iterations.
Our presentation is purely expository; we disavow any claim we might make
to originality --- in fact, our observations should be obvious to experts
in stochastic approximation. We opt for concision over full mathematical rigor,
and assume that the reader is familiar with the subjects reviewed
by~\citet{bach}.

The remainder of the present paper has the following structure:
Section~\ref{defs} sets mathematical notation
for the stochastic approximation/optimization considered in the sequel.
Section~\ref{acceleration} elaborates a principle for accelerating
the optimization. Section~\ref{minibatch} discusses minibatching.
Section~\ref{numerical} supplements the numerical examples
of~\citet{bach} and~\citet{sutskever-martens-dahl-hinton}.
Section~\ref{conclusion} reiterates the above, concluding the paper.

\section{A mathematical formulation}
\label{defs}

Machine learning commonly involves a cost/loss function $c$
and a single random sample $X$ of test data;
we then want to select parameters $\theta$ minimizing the expected value
\begin{equation}
\label{objective}
e(\theta) = \E[c(\theta,X)]
\end{equation}
($e$ is known as the ``error'' or ``risk'' --- specifically,
the ``test'' error, as opposed to ``training'' error).
The cost $c$ is generally a nonnegative scalar-valued function,
whereas $\theta$ and $X$ can be vectors of real or complex numbers.
In supervised learning,
$X$ is a vector which contains both ``input'' or ``feature'' entries
and corresponding ``output'' or ``target'' entries
--- for example, for a regression the input entries can be regressors
and the outputs can be regressands; for a classification the outputs
can be the labels of the correct classes.
For training, we have no access to $X$ (nor to $e$) directly,
but can only access independent and identically distributed (i.i.d.)\ samples
from the probability distribution $p$ from which $X$ arises.
Of course, $e(\theta)$ depends only on $p$, not on any particular sample drawn
from $p$.

To finish setting our notational conventions,
we define the deviation or dispersion
\begin{equation}
\label{ddef}
d(\theta,X) = c(\theta,X) - e(\theta),
\end{equation}
yielding the decomposition
\begin{equation}
\label{decomp}
c(\theta,X) = e(\theta) + d(\theta,X).
\end{equation}
Combining~(\ref{ddef}) and~(\ref{objective}) yields that
\begin{equation}
\E[d(\theta,X)] = 0
\end{equation}
for all $\theta$, that is, the mean of $d(\theta,X)$ is 0
(whereas the mean of $c(\theta,X)$ is $e(\theta)$).

\section{Accelerated learning}
\label{acceleration}

For definiteness, we limit consideration to algorithms which may access
only values of $c(\theta,x)$ and its first-order derivatives/gradient
with respect to $\theta$.
We assume that the cost $c$ is calibrated (shifted by an appropriate constant)
such that the absolute magnitude of $c$ is a good absolute gauge of the quality
of $\theta$ (lower cost is better, of course).
As reviewed, for example, by~\citet{bach},
the algorithm of Robbins and Monro (online stochastic gradient descent)
minimizes the test error $e$ in~(\ref{objective}) using a number of samples
(from the probability distribution $p$) that is optimal ---
optimal to within a constant factor which depends on the starting point
for $\theta$. 
Despite this kind of optimality, if $e$ happens to start much larger than
$d = c-e$, then a higher-order method (such as the Nesterov acceleration
discussed by~\citet{bach} and others) can dramatically reduce the constant.
We say that $\theta$ is ``poor'' to mean that $e(\theta)$ is much larger
than $d(\theta,X)$.
If $\theta$ is poor, that is,
\begin{equation}
e(\theta) \gg |d(\theta,X)|,
\end{equation}
then~(\ref{decomp}) yields that
\begin{equation}
c(\theta,X) = e(\theta) + d(\theta,X) \approx e(\theta)
\end{equation}
is effectively deterministic,
and so Nesterov and other methods from~\citet{bach} could accelerate
the optimization until the updated $\theta$ is no longer poor.

Clearly, $\theta$ being poor, that is,
$e(\theta)$ being much larger than $d(\theta,X)$, simply means that
$c(\theta,x) = e(\theta) + d(\theta,x)$ depends
mostly just on the probability distribution $p$ underlying $X$,
but not so much on any particular sample $x$ from $p$.
(For supervised learning, the probability distribution $p$ is joint
over the inputs and outputs, encoding the relation between the inputs
and their respective outputs.)
If $\theta$ is poor, that is, $e(\theta)$ is much larger than $d(\theta,X)$,
so that $c(\theta,X) \approx e(\theta)$ is mostly deterministic,
then Nesterov methods can be helpful.

Of course, optimization performance may suffer
from applying a higher-order method to a rough, random objective
(say, to a random variable $c(\theta,X)$ estimating $e(\theta)$
when the standard deviation of $c(\theta,X)$ is high relative to $e(\theta)$).
For instance, Nesterov methods essentially sum across several iterations;
if the values being summed were stochastically independent,
then the summation would actually increase the variance.
{\it In accordance with the optimality of the method of Robbins and Monro
(online stochastic gradient descent),
Nesterov acceleration can be effective only when the estimates
of the objective function are dominantly deterministic,
with the objective function being the test error $e$ from~(\ref{objective}).}

In practice, we can apply higher-order methods during the initial iterations
when the starting point is poor, gradually transitioning
to the original method of Robbins and Monro (stochastic gradient descent)
as the test error approaches its lower limit, that is, as stochastic variations
in estimates of the objective function and its derivatives become important.

\section{Minibatches}
\label{minibatch}

Some practical considerations detailed, for example,
by~\cite{lecun-bottou-bengio-haffner} are the following:

Many modern microprocessor architectures can leverage the parallelism inherent
in batch processing. With such parallel (and partly parallel) processors,
minimizing the number of samples drawn
from the probability distribution underlying the data
may not be the most efficient possibility.
Rather than updating the parameters being learned for random individual samples
as in online stochastic gradient descent,
the common current practice is to draw all at once a number of samples
--- a collection of samples known as a ``minibatch'' ---
and then update the parameters simultaneously
for all these samples before proceeding to the next minibatch.
With sufficient parallelism, processing an entire minibatch
may take little longer than processing a single sample.

Averaging the estimates of the objective function
(that is, of the test error $e$) and its derivatives over all samples
in a minibatch yields estimates with smaller standard deviations,
effectively making more parameter values be ``poor,''
so that Nesterov acceleration is more effective for more iterations.
Moreover, minibatches provide (essentially for free) estimates
of the standard deviations of the estimates of $e$.
The number of initial iterations for which Nesterov methods
can accelerate the optimization can be made arbitrarily large
by setting the size of the minibatches arbitrarily large.
That said, smaller minibatch sizes typically require fewer samples in total
to approach full convergence (albeit typically with more iterations,
processing one minibatch per iteration).
Also, after sufficiently many iterations,
stochastic variations in estimates of the objective function
and its derivatives can become important,
and continuing to use Nesterov acceleration in these later iterations
would be counterproductive.
Again, we recommend applying higher-order methods during the initial iterations
when the starting point is poor, noting that larger minibatches make
the higher-order methods advantageous for more iterations before the point
requiring turning off the Nesterov acceleration.

\section{Numerical experiments}
\label{numerical}

\begin{figure}
\caption{Test accuracies for various settings
of momentum (mom.)\ and learning rates (l.r.) [minibatch size 100]}
\label{mini100}

\begin{center}
(a)\parbox{.91\textwidth}{\includegraphics[width=.9\textwidth]{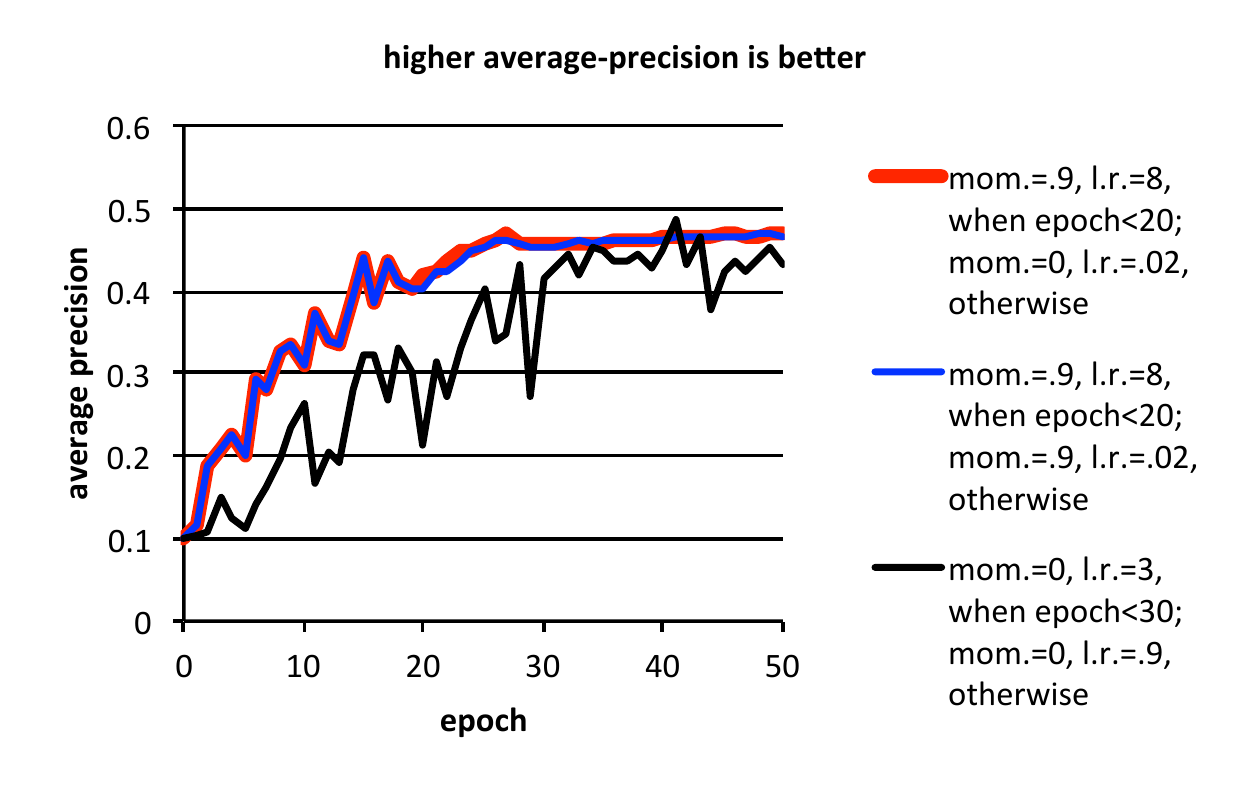}}

\vspace{.05in}

(b)\parbox{.91\textwidth}{\includegraphics[width=.9\textwidth]{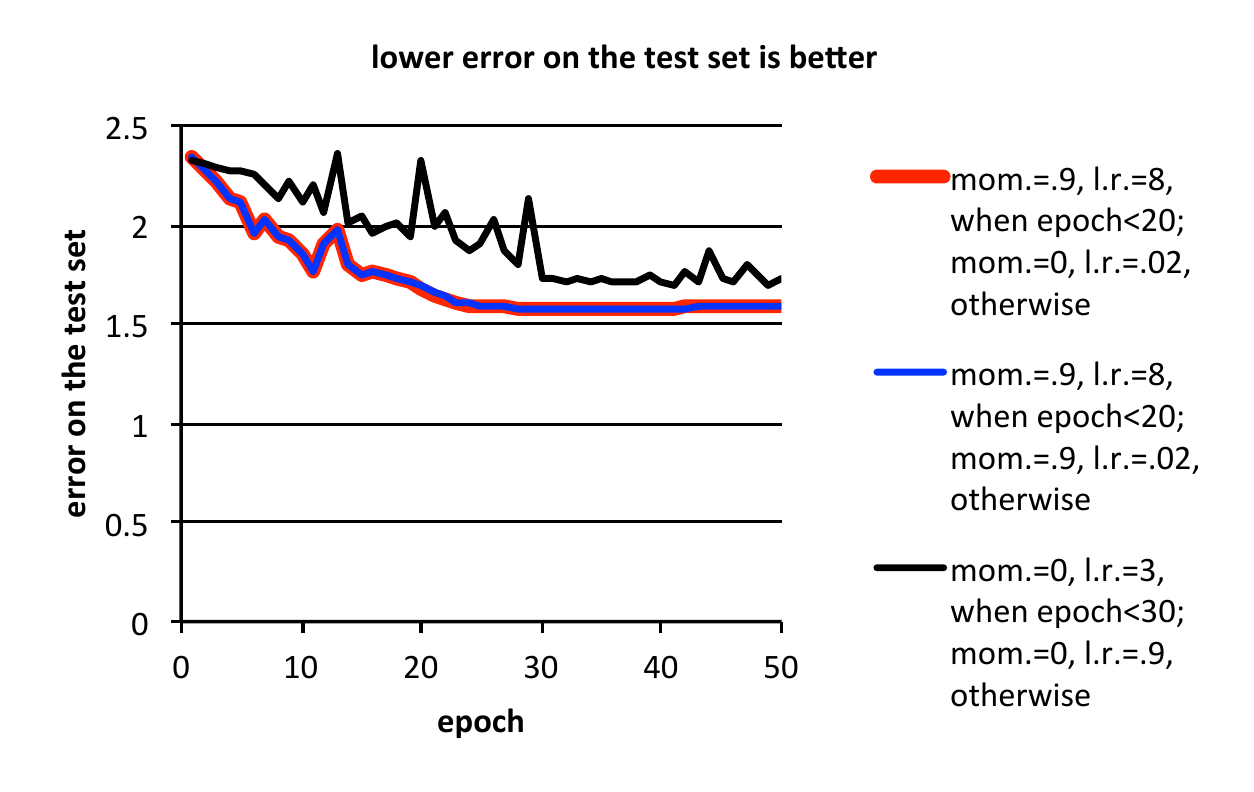}}
\end{center}
\end{figure}

\begin{figure}
\caption{Test accuracies for various settings
of momentum (mom.)\ and learning rates (l.r.) [minibatch size 1]}
\label{mini1}

\begin{center}
(a)\parbox{.91\textwidth}{\includegraphics[width=.9\textwidth]{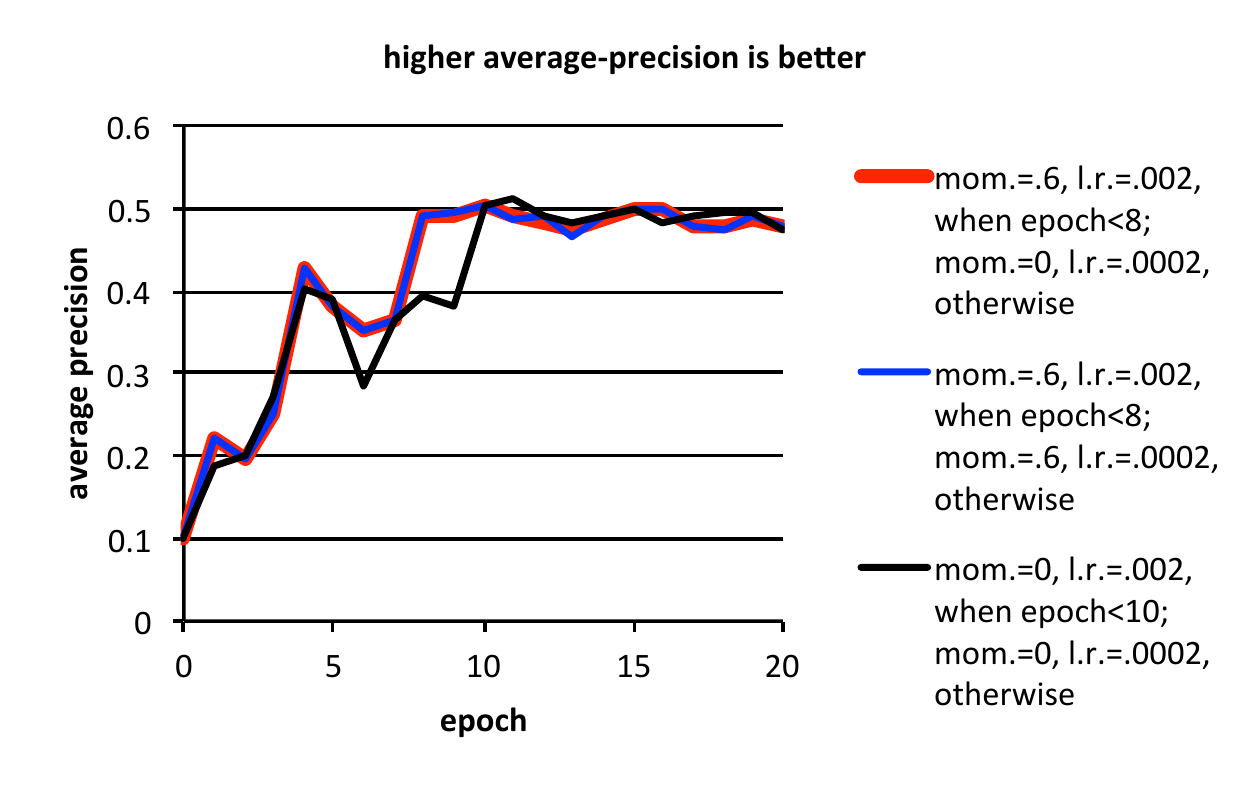}}

\vspace{.05in}

(b)\parbox{.91\textwidth}{\includegraphics[width=.9\textwidth]{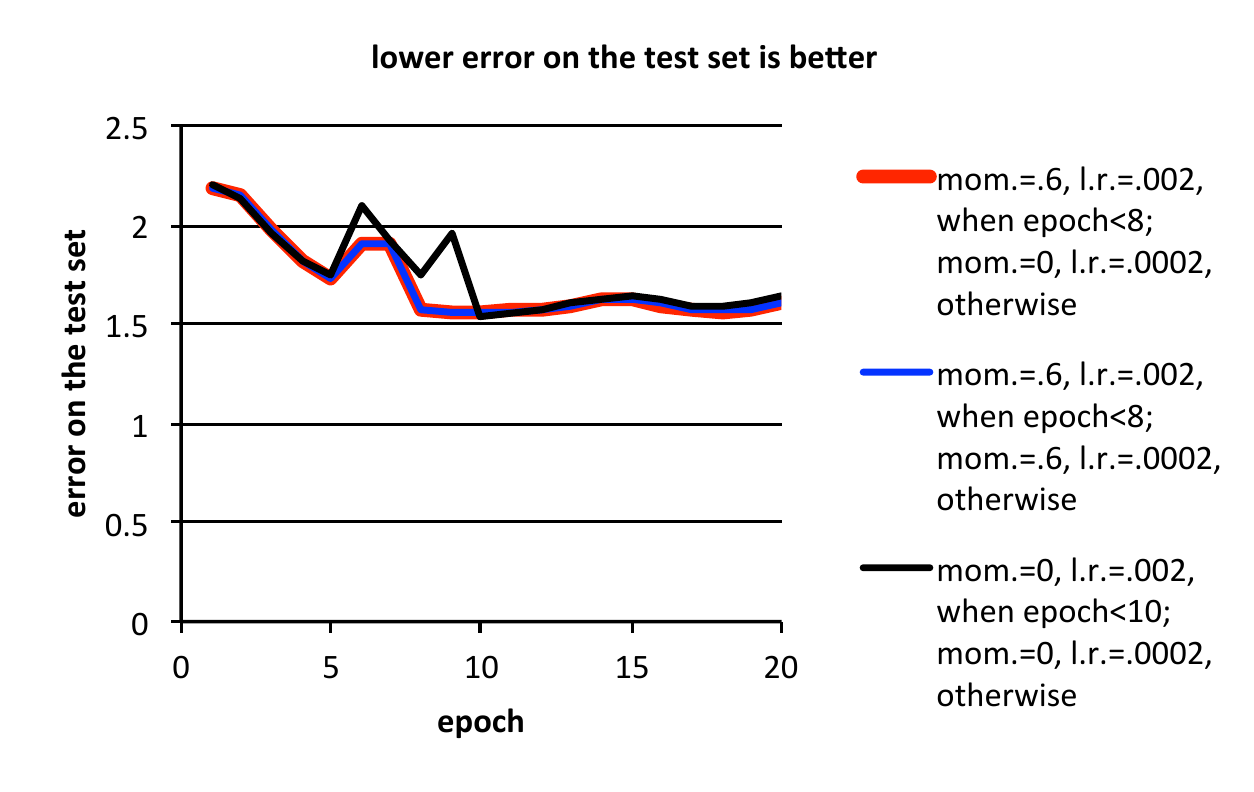}}
\end{center}
\end{figure}

The present section supplements the examples
of~\citet{bach} and~\citet{sutskever-martens-dahl-hinton}
with a few more experiments indicating that a higher-order method
--- namely, momentum, a form of Nesterov acceleration --- can help a bit
in training convolutional networks.
All training reported is via the method of Robbins and Monro
(stochastic gradient descent), with minibatches first
of size 1 sample per iteration (as in the original, online algorithm) and then
of size 100 samples per iteration
(size 100 is among the most common choices used in practice
on modern processors which can leverage the parallelism inherent
in batch processing).
Section~\ref{minibatch} and the third-to-last paragraph of the present section
describe the minibatching in more detail.
Figures~\ref{mini100} and~\ref{mini1} report the accuracies
for various configurations of ``momentum'' and ``learning rates'';
momentum appears to accelerate training somewhat,
especially with the larger size of minibatches.

Each iteration subtracts from the parameters being learned
a multiple of a stored vector, where the multiple is the ``learning rate,''
and the stored vector is the estimated gradient plus
the stored vector from the previous iteration,
with the latter stored vector scaled by the amount of ``momentum'':
\begin{equation}
\theta^{(i)} = \theta^{(i-1)} - \alpha^{(i)}_{\it l.r.} \, v^{(i)},
\end{equation}
with
\begin{equation}
v^{(i)} = \beta^{(i)}_{\it mom.} \, v^{(i-1)} + \frac{1}{k} \sum_{j=1}^k \left.
\left(\frac{\partial c}{\partial \theta}\left(\theta,x^{(i,j)}\right)\right)
\right|_{\theta=\theta^{(i-1)}},
\end{equation}
where $i$ is the index of the iteration,
$\theta^{(i)}$ is the updated vector of parameters,
$\theta^{(i-1)}$ is the previous vector of parameters,
$v^{(i)}$ is the updated stored auxiliary vector,
$v^{(i-1)}$ is the previous stored auxiliary vector,
$\alpha^{(i)}_{\it l.r.}$ is the learning rate,
$\beta^{(i)}_{\it mom.}$ is the amount of momentum,
$x^{(i,j)}$ is the $j$th training sample in the $i$th minibatch,
$k$ is the size of the minibatch,
and $\partial c / \partial \theta$ is the gradient of the cost $c$
with respect to the parameters $\theta$.

Following~\citet{lecun-bottou-bengio-haffner}
exactly as done by~\citet{chintala-ranzato-szlam-tian-tygert-zaremba},
the architecture for generating the feature activations
is a convolutional network (convnet) consisting of a series of stages,
with each stage feeding its output into the next.
The last stage has the form of a multinomial logistic regression,
applying a linear transformation to its inputs, followed by
the ``softmax'' detailed by~\citet{lecun-bottou-bengio-haffner}, thus
producing a probability distribution over the classes in the classification.
The cost/loss is the negative of the natural logarithm of the probability
so assigned to the correct class.
Each stage before the last convolves each image from its input
against several learned convolutional kernels, summing together
the convolved images from all the inputs into several output images,
then takes the absolute value of each pixel of each resulting image,
and finally averages over each patch in a partition of each image
into a grid of $2 \times 2$ patches.
All convolutions are complex valued and produce pixels
only where the original images cover all necessary inputs (that is,
a convolution reduces each dimension of the image by one less than the size
of the convolutional kernel).
We subtract the mean of the pixel values from each input image
before processing with the convnet, and we append
an additional feature activation feeding into the ``softmax''
to those obtained from the convnet,
namely the standard deviation of the set of values of the pixels in the image.

The data is a subset of the 2012 ImageNet set of~\citet{imagenet2012},
retaining 10 classes of images, representing each class by 100 samples
in a training set and 50 per class in a testing set.
Restricting to this subset facilitated more extensive experimentation
(optimizing hyperparameters more extensively, for example).
The images are full color, with three color channels.
We neither augmented the input data nor regularized the cost/loss functions.
We used the Torch7 platform --- http://torch.ch --- for all computations.

Table~\ref{tab} details the convnet architecture we tested.
``Stage'' specifies the positions of the indicated layers in the convnet.
``Input channels'' specifies the number of images input to the given stage
for each sample from the data.
``Output channels'' specifies the number of images output from the given stage.
Each input image is convolved against a separate, learned convolutional kernel
for each output image (with the results of all these convolutions
summed together for each output image).
``Kernel size'' specifies the size of the square grid of pixels
used in the convolutions.
``Input channel size'' specifies the size of the square grid of pixels
constituting each input image.
``Output channel size'' specifies the size of the square grid of pixels
constituting each output image.
The feature activations that the convnet produces
feed into a linear transformation followed by a ``softmax,''
as detailed by~\citet{lecun-bottou-bengio-haffner}.

For the minibatch size 100, rather than updating the parameters
being learned for randomly selected individual images from the training set
as in online stochastic gradient descent, we instead do the following:
we randomly permute the training set and partition this permuted set
of images into subsets of 100, updating the parameters simultaneously
for all 100 images constituting each of the subsets (known as ``minibatches''),
processing the series of minibatches in series.
Each sweep through the entire training set is known as an ``epoch.''
\citet{lecun-bottou-bengio-haffner}, among others, made the above terminology
the standard for training convnets.
The horizontal axes in the figures count the number of epochs.

\begin{figure}
\caption{Estimated coefficients of variation (CVs)\ of the cost $c(\theta,X)$
in~(\ref{objective}) and~(\ref{decomp}), using all samples in each minibatch
to construct estimates of the CVs (the estimates of the CVs are
themselves subject to stochastic variations;
the distribution of the plotted points within an epoch provides an indication
of the probability distribution of the estimates of the CVs)
[minibatch size 100]}
\label{randomness}

\begin{center}
\parbox{.91\textwidth}{\includegraphics[width=.9\textwidth]{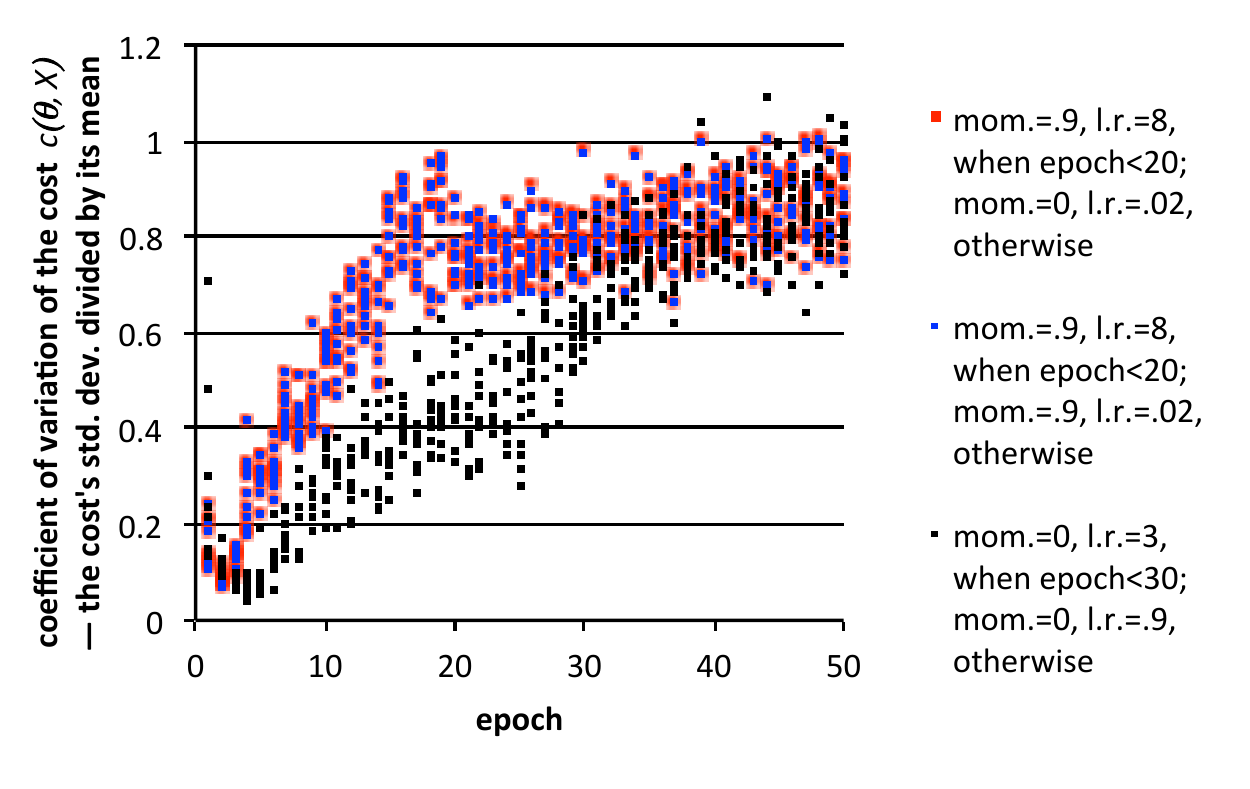}}
\end{center}
\end{figure}

Figures~\ref{mini100} and~\ref{randomness} present the results for minibatches
of size 100 samples per iteration;
Figure~\ref{mini1} presents the results for minibatches of size 1 sample
per iteration.
``Average precision'' is the fraction of all classifications
which are correct, choosing only one class for each input sample image
from the test set.
``Error on the test set'' is the average over all samples in the test set
of the negative of the natural logarithm of the probability assigned
to the correct class (assigned by the ``softmax'').
``Coefficient of variation'' is an estimate of the standard deviation
of $c(\theta,X)$ (from~(\ref{objective})) divided by the mean of~$c(\theta,X)$,
that is, an estimate of the standard deviation of $d(\theta,X)$
(from~(\ref{ddef})) divided by $e(\theta)$
(from~(\ref{objective}) and~(\ref{ddef})).
Rolling off momentum as the coefficient of variation increases
can be a good idea.

Please beware that these experiments are far from definitive,
and even here the gains from using momentum seem to be marginal.
Even so, as Section~\ref{minibatch} discusses,
minibatching effectively reduces the standard deviations
of estimates of the objective function $e$,
making Nesterov acceleration more effective for more iterations.

\section{Conclusion}
\label{conclusion}

Though in many settings the method of Robbins and Monro
(stochastic gradient descent) is optimal for good starting points,
higher-order methods (such as momentum and Nesterov acceleration) can help
during early iterations of the optimization when the parameters being optimized
are poor in the sense discussed above.
The opportunity for accelerating the optimization is clear theoretically
and apparently observable via numerical experiments.
Minibatching makes the higher-order methods advantageous for more iterations.
That said, higher order and higher accuracy need not be the same,
as higher order guarantees only that accuracy increase at a faster rate
--- higher order guarantees higher accuracy only after enough iterations.

\section*{Acknowledgements}

We would like to thank Francis Bach, L\'eon Bottou, Yann Dauphin,
Piotr Doll\'ar, Yann LeCun, Marc'Aurelio Ranzato, Arthur Szlam, and Rachel Ward.

\appendix
\section{A simple analytical example}

This appendix works through a very simple one-dimensional example.

Consider the cost $c$ as a function of a real scalar parameter $\theta$ and
sample of data $x$, defined via
\begin{equation}
\label{simpex}
c(\theta,x) = (\theta-x)^2.
\end{equation}

Suppose that $X$ is the (Rademacher) random variable taking the value $1$
with probability $1/2$ and the value $-1$ with probability $1/2$.
Then, the coefficient of variation of $c(\theta,X)$, that is,
the standard deviation $\sigma(\theta)$ of $c(\theta,X)$
divided by the mean $e(\theta)$ of $c(\theta,X)$, is
\begin{equation}
\label{CV}
\frac{\sigma(\theta)}{e(\theta)} = \frac{2|\theta|}{\theta^2+1}:
\end{equation}

\bigskip

\noindent {\it (proof)}
$X$ taking the value $1$ with probability $1/2$ and the value $-1$
with probability $1/2$ yields that
\begin{equation}
\label{normalized}
|X| = 1
\end{equation}
and that the expected value of $X$ is 0,
\begin{equation}
\label{centered}
\E[X] = 0.
\end{equation}
Combining~(\ref{simpex}) and~(\ref{normalized}) then yields
\begin{equation}
\label{simpler}
c(\theta,X) = \theta^2 + 1 - 2\theta X.
\end{equation}
Moreover, together with the definitions~(\ref{objective}) and~(\ref{ddef}),
combining~(\ref{simpler}) and~(\ref{centered}) yields
\begin{equation}
\label{eex}
e(\theta) = \E[c(\theta,X)] = \theta^2 + 1
\end{equation}
and
\begin{equation}
\label{dex}
d(\theta,X) = c(\theta,X) - e(\theta) = -2\theta X.
\end{equation}
The mean of $c(\theta,X)$ is this $e(\theta)$ in~(\ref{eex}).
The standard deviation of $c(\theta,X)$ is the same as the standard deviation
of~$d(\theta,X)$ in~(\ref{dex}), which is just
\begin{equation}
\label{stddev}
\sigma(\theta) = 2|\theta|,
\end{equation}
due to~(\ref{dex}) combined
with $X$ taking the value $1$ with probability $1/2$ and the value $-1$
with probability $1/2$.
Combining~(\ref{eex}) and~(\ref{stddev}) yields~(\ref{CV}), as desired.
\hfill $\blacksquare$

\bigskip

The secant method (a higher-order method) consists of the iterations
updating $\theta^{(i-2)}$ and~$\theta^{(i-1)}$ to~$\theta^{(i)}$ via
\begin{multline}
\label{secant}
\theta^{(i)} = \theta^{(i-1)}
- \frac{\partial c}{\partial \theta}\left(\theta^{(i-1)},x^{(i-1)}\right)
\\ \cdot \left(\theta^{(i-1)}-\theta^{(i-2)}\right) \bigg/
\left(\frac{\partial c}{\partial \theta}\left(\theta^{(i-1)},x^{(i-1)}\right)
-\frac{\partial c}{\partial \theta}\left(\theta^{(i-2)},x^{(i-2)}\right)\right),
\end{multline}
where $x^{(i-2)}$,~$x^{(i-1)}$, and~$x^{(i)}$ are independent realizations
(random samples) of $X$ and the derivative of $c(\theta,x)$ from~(\ref{simpex})
with respect to $\theta$ is
\begin{equation}
\label{gradex}
\frac{\partial c}{\partial \theta} = 2(\theta-x);
\end{equation}
combining~(\ref{secant}) and~(\ref{gradex}) and simplifying yields
\begin{equation}
\label{explicit}
\theta^{(i)}
= \left( \theta^{(i-1)} x^{(i-2)} - \theta^{(i-2)} x^{(i-1)} \right) \bigg/
\left(\theta^{(i-1)}-x^{(i-1)}-\theta^{(i-2)}+x^{(i-2)}\right).
\end{equation}
In the limit that both $x^{(i-1)}$ and~$x^{(i-2)}$ be 0,
(\ref{explicit}) shows that $\theta^{(i)}$ is also 0
(provided that $\theta^{(i-1)}$ and~$\theta^{(i-2)}$ are distinct), that is,
the secant method finds the optimal value for $\theta$ in a single iteration
in such a limit.
In fact, (\ref{eex}) makes clear that the optimal value for $\theta$ is 0,
while combining~(\ref{explicit}) and~(\ref{normalized}) yields that
\begin{equation}
|\theta^{(i)}| \le \left( |\theta^{(i-1)}| + |\theta^{(i-2)}| \right)
\bigg/ \left( |\theta^{(i-1)}-\theta^{(i-2)}| - 2 \right),
\end{equation}
so that $|\theta^{(i)}|$ is likely to be reasonably small (around 1 or so)
even if $|\theta^{(i-1)}|$ or $|\theta^{(i-2)}|$
(or both) are very large and random and $|X| = 1$.

However, these iterations generally fail to converge
to any value much smaller than unit magnitude: with probability $1/2$,
indeed, $x^{(i-2)} = x^{(i-1)} = 1$ or $x^{(i-2)} = x^{(i-1)} = -1$;
hence (\ref{explicit}) yields that $|\theta^{(i)}| = 1$
with probability at least $1/2$,
assuming that $\theta^{(i-2)}$ and $\theta^{(i-1)}$ are distinct
(while in the degenerate case that $\theta^{(i-2)}$ and $\theta^{(i-1)}$
take exactly the same value, $\theta^{(i)}$ also takes that same value).

All in all, a sensible strategy is to start with a higher-order method
(such as the secant method) when $|\theta|$ is large,
and transition to the asymptotically optimal method of Robbins and Monro
(stochastic gradient descent) as $|\theta|$ becomes of roughly unit magnitude.
The transition can be based on $|\theta|$ or
on estimates of the coefficient of variation
--- due to~(\ref{CV}), the coefficient of variation
is essentially inversely proportional to $|\theta|$ when $|\theta|$ is large.

\begin{table}
\caption{Architecture of the convolutional network}
\label{tab}
\vspace{.5em}
\begin{center}
\begin{tabular}{cccccc}
       & input    & output   & kernel & input        & output       \\
 stage & channels & channels & size   & channel size & channel size \\\hline
 first &   3 &  16 & $5 \times 5$ & $128 \times 128$ & $62 \times 62$ \\
second &  16 &  64 & $3 \times 3$ &   $62 \times 62$ & $30 \times 30$ \\
 third &  64 & 256 & $3 \times 3$ &   $30 \times 30$ & $14 \times 14$ \\
fourth & 256 & 256 & $3 \times 3$ &   $14 \times 14$ &   $6 \times 6$
\end{tabular}
\end{center}
\end{table}

\clearpage

\bibliography{arxiv}
\bibliographystyle{apalike}

\end{document}